\documentclass[letterpaper]{article} 
\usepackage{aaai23}  
\usepackage{times}  
\usepackage{helvet}  
\usepackage{courier}  
\usepackage[hyphens]{url}  
\usepackage{graphicx} 
\usepackage{natbib}  
\usepackage{caption} 
\usepackage{algorithm}
\usepackage{algorithmic}
\usepackage{tablefootnote}

\usepackage{newfloat}
\usepackage{listings}
\DeclareCaptionStyle{ruled}{labelfont=normalfont,labelsep=colon,strut=off} 
\floatstyle{ruled}
\newfloat{listing}{tb}{lst}{}
\floatname{listing}{Listing}
\usepackage{algorithm}
\usepackage{algorithmic}

\usepackage{url}
\usepackage{amsmath}
\usepackage{caption}
\usepackage{subcaption}
\usepackage{amsmath,amsfonts,amssymb}
\usepackage{mathrsfs}
\usepackage{multirow}
\usepackage{makecell}
\usepackage{colortbl}
\usepackage{booktabs}
\usepackage{array}
\usepackage{newfloat}
\usepackage{listings}
\definecolor{lightgray}{rgb}{0.9, 0.9, 0.9}
\definecolor{ballblue}{rgb}{0.13, 0.67, 0.8}
\usepackage[colorlinks,citecolor=ballblue]{hyperref}
\newcommand{\STAB}[1]{\begin{tabular}{@{}c@{}}#1\end{tabular}}

\title{Convolutional Bypasses Are Better Vision Transformer Adapters}

\author {
    Shibo Jie,
    Zhi-Hong Deng
}
\affiliations {
    School of Artificial Intelligence, Peking University\\

    parsley@pku.edu.cn, zhdeng@pku.edu.cn
}

\usepackage{bibentry}

\begin{document}

\maketitle

\begin{abstract}
The pretrain-then-finetune paradigm has been widely adopted in computer vision. But as the size of \emph{Vision Transformer} (ViT) grows exponentially, the full finetuning becomes prohibitive in view of the heavier storage overhead. Motivated by \emph{parameter-efficient transfer learning} (PETL) on language transformers, recent studies attempt to insert lightweight adaptation modules (e.g., adapter layers or prompt tokens) to pretrained ViT and only finetune these modules while the pretrained weights are frozen. However, these modules were originally proposed to finetune language models and did not take into account the prior knowledge specifically for visual tasks. In this paper, we propose to construct \emph{Convolutional Bypasses} (Convpass) in ViT as adaptation modules, introducing only a small amount (less than 0.5\% of model parameters) of trainable parameters to adapt the large ViT. Different from other PETL methods, Convpass benefits from the hard-coded inductive bias of convolutional layers and thus is more suitable for visual tasks, especially in the low-data regime. Experimental results on VTAB-1K benchmark and few-shot learning datasets show that Convpass outperforms current language-oriented adaptation modules, demonstrating the necessity to tailor vision-oriented adaptation modules for adapting vision models.

\end{abstract}

\section{Introduction}

\begin{figure}[t]
     \centering
         \includegraphics[width=0.4\textwidth]{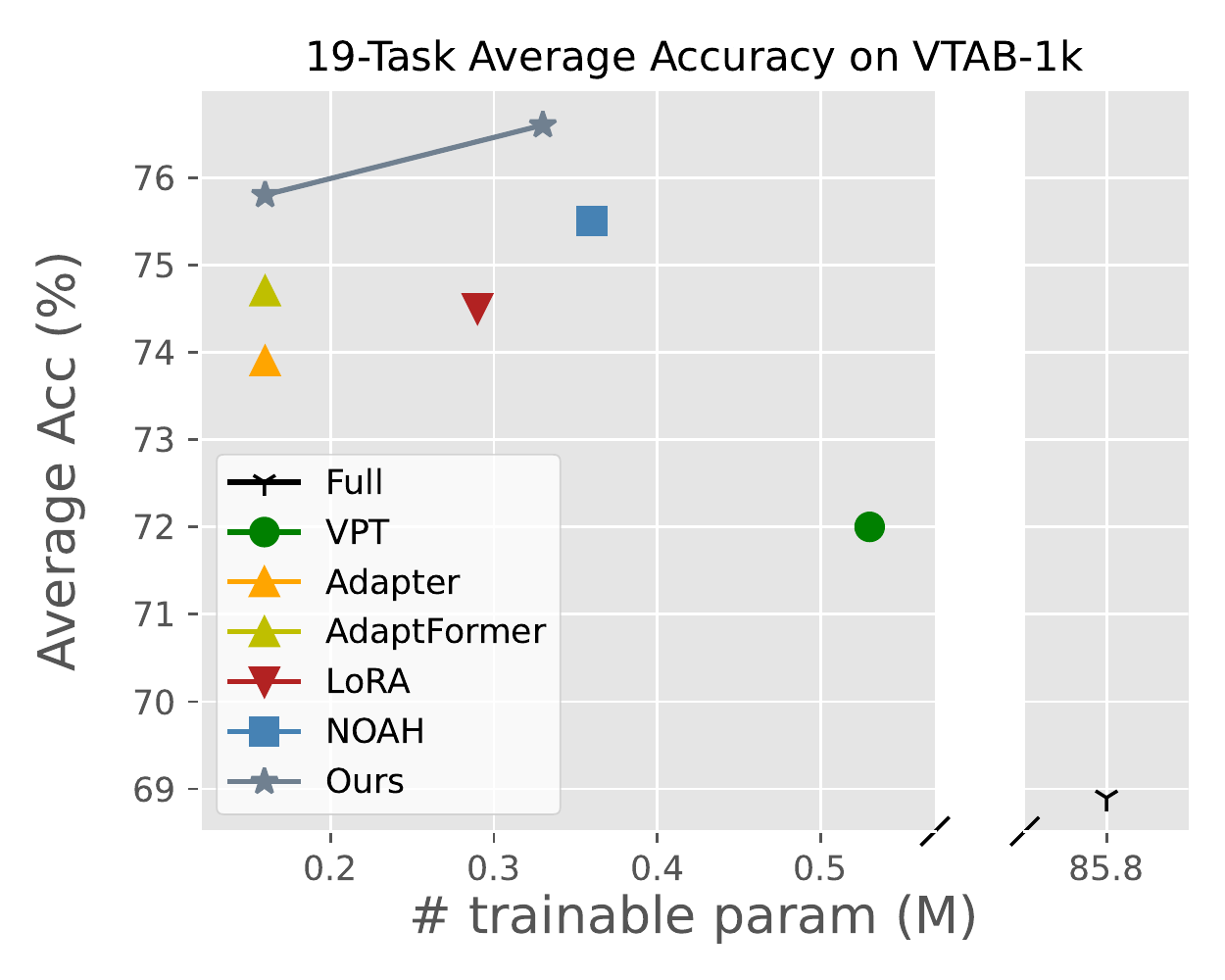}
         \caption{\textbf{Average accuracy vs. number of trainable parameters on VTAB-1K benchmark.} Our vision-oriented Convpass outperforms other language-oriented methods.}
         \label{fig:intro}
\end{figure}
Pretraining on large-scale datasets (e.g., ImageNet) and then fully finetuning on downstream tasks has become the de-facto paradigm to achieve state-of-the-art (SOTA) performance on visual tasks~\cite{bit}. However, this paradigm is not storage-efficient -- it requires one to store a whole model for each downstream task. Recently, as \emph{Vision Transformer} (ViT)~\cite{vit} dominates vision field gradually, the size of vision models has grown exponentially (58M of ResNet-152~\cite{resnet} vs. 1843M of ViT-G~\cite{vitg}), which creates the demand for \emph{parameter-efficient transfer learning} (PETL) on ViT. 

Fortunately, since transformer was first adopted in \emph{neural language processing} (NLP)~\cite{transformer}, PETL on large pretrained language models has been studied sufficiently~\cite{adapter,lora,prefix,towards}, which can be easily ported to ViT. Concretely, these PETL methods insert lightweight adaptation modules into the pertrained models, freeze the pretrained weights, and finetune these modules end-to-end to adapt to downstream tasks. Recent work has verified the effectiveness of these PETL methods on ViT~\cite{vpt,noah}, but we raise a question: \emph{Are these modules designed for the language models optimal for vision models as well?} 

It is known that NLP tasks and visual tasks desire different inductive bias, which profoundly affects the model architecture design. By analyzing current PETL methods from an unraveled perspective, we argue that these methods, called ``language-oriented modules'', also imply the inductive bias for language, e.g., weak spatial relation and support for variable-length input. Therefore, a better adaptation module for ViT should also reflect visual inductive bias, such as spatial locality and 2D neighborhood structure, which is referred to as ``vision-oriented modules''.

When a model (e.g., ViT) has weak inductive bias, it needs a large amount of data to learn the inductive bias from scratch. This may not be a serious problem in the pretraining process, since we can leverage easily accessible unlabeled data for self-supervised learning~\cite{beit,mae}, or resort to multi-modal pretraining~\cite{clip,coca}. However, data of downstream tasks is usually collected from specific domains that may be expensive or hard to acquire. Therefore, besides the inductive bias learned from pretraining data, a well-designed vision-oriented PETL module is expected to introduce additional inductive bias and improve data efficiency much further.


In this paper, we propose to construct \emph{Convolutional Bypasses} (Convpass) in ViT as adaptation modules. Convpass is an inserted convolutional bottleneck block  parallel to the MHSA or MLP block, which ``bypasses'' the original ViT block. It reconstructs the spatial structure of the token sequence and performs convolution on image tokens and \texttt{[cls]} token individually. During finetuning, only these Convpass modules and the classification head are updated. Due to the hard-coded locality of convolutional layers, Convpass can capture visual information more efficiently, especially when the downstream data is limited. As shown in Figure~\ref{fig:intro}, Convpass only introduces and tunes about 0.33M new parameters for a ViT-B of 86M, while achieving better performance than both full finetuning and current SOTA language-oriented methods on 19-task VTAB benchmark~\cite{vtab}. Further experiments on few-shot learning demonstrate that Convpass also outperforms other baselines in the low-data regime, and can be directly used on vision-language model~\cite{clip} with good domain generalization performance.

We summarize the contributions as follows:
\begin{itemize}
    \item We point out the weak visual inductive bias of current PETL methods that limits their performance on ViT.
    \item We propose Convpass, a simple yet effective PETL method which leverages trainable convolutional blocks as bypasses to adapt pretrained ViT to downstream visual tasks.
    \item Experimental results show that Convpass outperforms previous language-oriented methods, indicating the necessity to tailor vision-oriented adaptation modules for vision models.
\end{itemize}

\section{Related Work}
\subsection{Vision Transformer}
Transformer-based models have achieved great success in NLP~\cite{bert,t5,gpt3}. ViT adopts this architecture in visual tasks by partitioning the images into patches which are embedded and flattened into 1D token sequences. 

In ViT, each layer consists of two kinds of blocks: \emph{Multi-Head Self-Attention} (MHSA) and \emph{Multi-Layer Perceptron} (MLP). In an MHSA block, the input sequence $\boldsymbol{X}\in \mathbb{R}^{N\times d}$ is firstly projected to query $\boldsymbol{Q}=\boldsymbol{XW}_q$, key $ \boldsymbol{K}=\boldsymbol{XW}_k$, and value $\boldsymbol{V}=\boldsymbol{XW}_v$, respectively, in which $\boldsymbol{W}_{q/k/v}\in \mathbb{R}^{d\times d}$. They are further divided into $N_h$ heads: $\{\boldsymbol{Q}^{(i)}\}^{N_h}_{i=1}, \{\boldsymbol{K}^{(i)}\}^{N_h}_{i=1}, \{\boldsymbol{V}^{(i)}\}^{N_h}_{i=1}$.
Then, the self-attention of a single head is  formulated as
\begin{equation*}
\begin{aligned}
    \textit{Attn-Head}^{(i)}(\boldsymbol{X})=
    \textit{Softmax}\left(\frac{\boldsymbol{Q}^{(i)}{\boldsymbol{K}^{(i)}}^\intercal}{\sqrt{d}}\right)\boldsymbol{V}^{(i)}
\end{aligned}
\end{equation*}
The outputs of all heads are further concatenated and linearly projected as the outputs of the MHSA block.

An MLP block consists of two fully-connected (FC) layers, whose weights are $\boldsymbol{W}_{1}\in \mathbb{R}^{d\times D}$ and $\boldsymbol{W}_{2}\in \mathbb{R}^{D\times d}$, respectively. Ignoring the bias parameters for simplicity, the MLP is  formulated as
\begin{equation*}
\begin{aligned}
    \textit{MLP}(\boldsymbol{X})=\textit{GELU}(\boldsymbol{X}\boldsymbol{W}_{1})\boldsymbol{W}_{2}
\end{aligned}
\end{equation*}

Since ViT has much less visual inductive bias, it performs worse than its convolutional counterparts (e.g., ResNet) when the training data is not sufficient. For this reason, some recent work proposes to introduce visual inductive bias into ViT~\cite{swin,cvt}, which significantly reduces its dependency on scale of dataset. However, vanilla ViT still has some nonnegligible advantages. Since vanilla ViT shares the same backbone as the transformer-based language models, it can leverage current SOTA multi-modal pretraining methods with a vast amount of auto-annotated image-text pairs~\cite{vlmo,coca}. Therefore, we still focus on PETL on vanilla ViT architecture, but propose to introduce hard-coded inductive bias by adaptation modules during finetuning instead of pretraining.

\subsection{Parameter-Efficient Transfer Learning}
PETL aims at using a small number of trainable parameters to adapt large models to downstream tasks. We here introduce some common PETL methods used for ViT. 

\textbf{Adapter}~\cite{adapter,adapterp} is a bottleneck MLP block composed of two fully connected layers, whose weights are $\boldsymbol{W}_{down}\in \mathbb{R}^{d\times h}$ and $\boldsymbol{W}_{up}\in \mathbb{R}^{h\times d}$, where $h << d$. Adapters are inserted into networks as residual connections, i.e., given an input $\boldsymbol{X}\in \mathbb{R}^{N\times d}$, the computation is  formulated as
$$\boldsymbol{X}'\leftarrow\boldsymbol{X}+\phi(\boldsymbol{X}\boldsymbol{W}_{down})\boldsymbol{W}_{up}$$
where $\phi$ is activation function such as GELU.

\citet{adapterp} propose to place Adapters after the MLP blocks (i.e., $\boldsymbol{X}$ is the output of MLP blocks), which has been proved to be an efficient design in previous literature~\cite{lora}, so we follow this setting in this paper. Besides the above design, \citet{towards} and \citet{adaptformer} also propose a parallel Adapter to adapt MLP blocks, which is formulated as
$$\boldsymbol{X}'\leftarrow\boldsymbol{X}+\textit{MLP}(\boldsymbol{X})+s\cdot\phi(\boldsymbol{X}\boldsymbol{W}_{down})\boldsymbol{W}_{up}$$
where $s$ is a hyperparameter, $\boldsymbol{X}$ is the input of MLP blocks. This Adapter design is referred to as \textbf{AdaptFormer} by \citet{adaptformer}.

\textbf{LoRA}~\cite{lora} learns the low-rank approximation of increments of $\boldsymbol{W}_{q}$ and $\boldsymbol{W}_{v}$. Formally, it decomposes $\Delta\boldsymbol{W}_{q/v}$ into $\boldsymbol{A}_{q/v}\boldsymbol{B}_{q/v}$, where $\boldsymbol{A}_{q/v}\in \mathbb{R}^{d\times r}, \boldsymbol{B}_{q/v}\in \mathbb{R}^{r\times d}$ and $r << d$. The query and key are computed as 
$$\boldsymbol{Q/K}=\boldsymbol{XW}_{q/k}+s\cdot\boldsymbol{XA}_{q/k}\boldsymbol{B}_{q/k}$$
in which $s$ is a scaling hyperparameter.

\textbf{VPT}~\cite{vpt} has a similar idea with P-Tuning v2~\cite{ptuningv2}. It concatenates the input $\boldsymbol{X}$ with several trainable prompts $\boldsymbol{P}\in \mathbb{R}^{l\times d}$ before each layer. This extended sequence is formulated as
$$\boldsymbol{X}'\leftarrow[\boldsymbol{X}, \boldsymbol{P}]$$
These prompts are then cut away at the end of a layer, and the prompts for the next layer are concatenated.

\textbf{NOAH}~\cite{noah} is a newly proposed PETL method for ViT, which combines the above three modules together and performs neural architecture search on hidden dimension $h$ of Adapter, rank $r$ of LoRA, and prompt length $l$ of VPT.

Note that although VPT and NOAH are proposed for visual tasks, their components are ported from NLP in essence. Therefore, all the aforementioned PETL methods can be classified as language-oriented methods. Other PETL methods such as \textbf{BitFit}~\cite{bitfit}, which finetunes the bias parameters only; and \textbf{Sidetune}~\cite{sidetune}, which finetunes a small side-network and interpolates between pretrained and side-tuned features, have been proved to perform rather poorly on ViT by \citet{vpt}.

\section{Methodology}
\subsection{Rethinking Adapters from an Unraveled View}
\begin{figure}[t]
     \centering
         \includegraphics[width=0.45\textwidth]{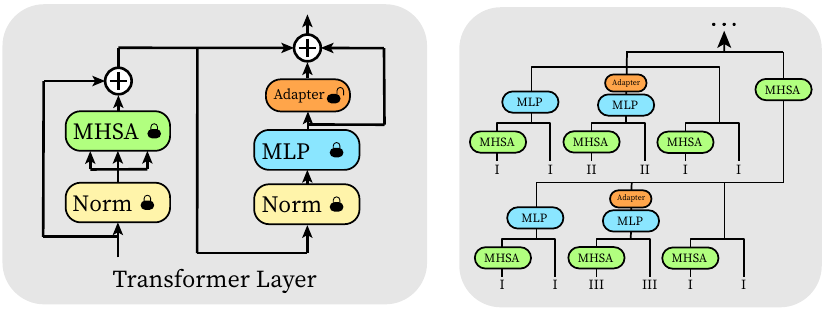}
         \caption{\textbf{Illustration of the unraveled view of ViT equipped with Adapter.} For simplicity, we show the unraveled view of a fragment of ViT (MHSA-MLP-MHSA) and the type of each path. Normalization layers are omitted.}
         \label{fig:unravel}
\end{figure}
Since Adapters and MHSA/MLP blocks all contain skip connections, we can unravel the ViT and rewrite it as a collection of paths. \citet{unravel} point out that the original network is an ensemble of unraveled paths, so we here give a look at these paths to analyze the property of the original network. 

As shown in Figure~\ref{fig:unravel}, a ViT equipped with Adapter can be viewed as an ensemble of three types of paths: (\textbf{Type I}) Frozen paths, which only contain MHSA/MLP blocks of the ViT. These paths are not trainable, and the sum of their outputs is identically equal to the output of the pretrained ViT. (\textbf{Type II}) MHSA-Adapter paths, where all MHSA blocks come before the first Adapter. (\textbf{Type III}) Adapter-MHSA paths, where at least one MHSA block is placed after an Adapter. 

Finetuning the Adapters is equivalent to fitting the changes of outputs by the paths of \textbf{Type II \& III}. In \textbf{Type II} paths, given the same input, the output tokens of the last MHSA blocks are unchanged, and there is no information exchange between tokens after that. Therefore, only the \textbf{Type III} paths, in fact, make changes to the token mixer of the pretrained ViT.

In a \textbf{Type III} paths, we can treat all Adapters and MLP blocks before an MHSA block as a part of its query/key/value transformation, i.e., complicate these transformations form linear mapping to
$$\boldsymbol{Q/K/V} = f_{q/k/v}(\boldsymbol{X})$$
where $f_{q/k/v}$ are channel-wise MLPs. Therefore, finetuning \textbf{Type III} paths can be considered as finetuning the MHSA with the complicated query/key/value transformations.

Meanwhile, since LoRA finetunes $\boldsymbol{W}_{q/v}$ in a low-rank subspace and VPT can be regarded as parallel and gated Adapters~\cite{towards}, all these language-oriented methods rely on tuning MHSA to adjust the token mixer on downstream tasks. MHSA, however, lacks visual inductive bias, which may perform poorly when the data of downstream visual tasks is limited. 

\begin{figure}[t]
     \centering
         \includegraphics[width=0.45\textwidth]{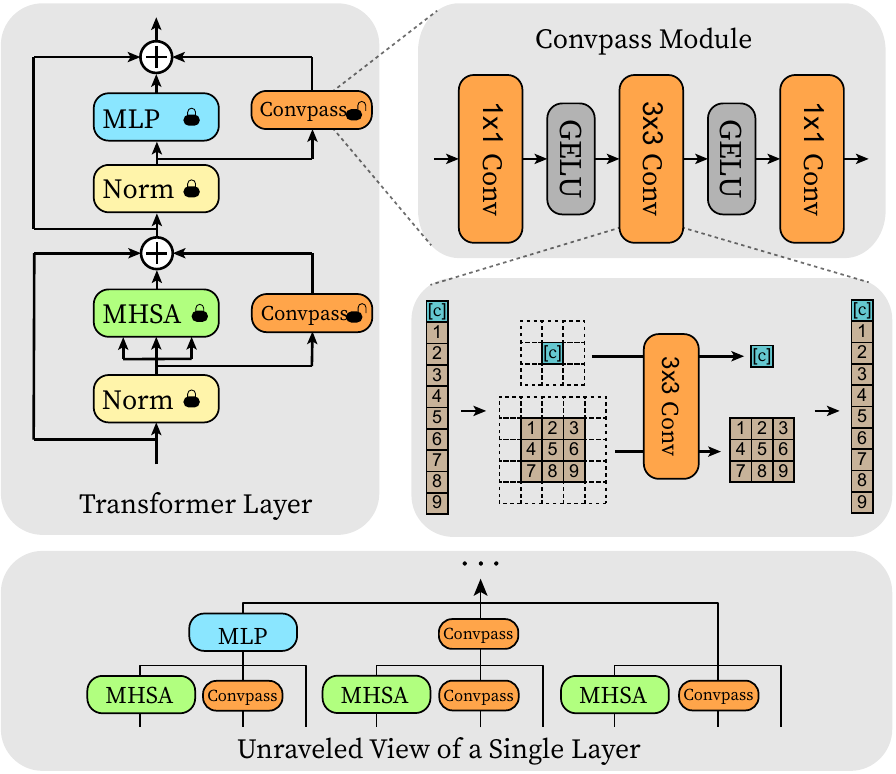}
         \caption{\textbf{Overview of the proposed method.} We restore the spatial structure of the token sequence, and use trainable ResNet-style convolutional blocks as bypasses. The \texttt{[cls]} token is regarded as an individual image.}
         \label{fig:cob}
\end{figure}

\begin{table*}[t]

\centering
\setlength{\tabcolsep}{0.3pt}
\scalebox{0.95}{
\begin{tabular}{p{2.2cm}<{}p{0.75cm}<{\centering}|p{0.75cm}<{\centering}p{0.75cm}<{\centering}p{0.75cm}<{\centering}p{0.75cm}<{\centering}p{0.75cm}<{\centering}p{0.75cm}<{\centering}p{0.75cm}<{\centering}|p{0.75cm}<{\centering}p{0.75cm}<{\centering}p{0.75cm}<{\centering}p{0.75cm}<{\centering}|p{0.75cm}<{\centering}p{0.75cm}<{\centering}p{0.75cm}<{\centering}p{0.75cm}<{\centering}p{0.75cm}<{\centering}p{0.75cm}<{\centering}p{0.75cm}<{\centering}p{0.75cm}<{\centering}|p{0.75cm}<{\centering}}
\toprule[1.5pt]
\multicolumn{2}{c|}{}&\multicolumn{7}{c|}{\textbf{Natural}}&\multicolumn{4}{c|}{\textbf{Specialized}}&\multicolumn{8}{c|}{\textbf{Structured}}&\\
&\multicolumn{1}{c|}{\STAB{\rotatebox[origin=c]{90}{\# param (M)}}}
&\multicolumn{1}{c}{\STAB{\rotatebox[origin=c]{90}{Cifar100}}}
&\multicolumn{1}{c}{\STAB{\rotatebox[origin=c]{90}{Caltech101}}}
&\multicolumn{1}{c}{\STAB{\rotatebox[origin=c]{90}{DTD}}}
&\multicolumn{1}{c}{\STAB{\rotatebox[origin=c]{90}{Flower102}}}
&\multicolumn{1}{c}{\STAB{\rotatebox[origin=c]{90}{Pets}}}
&\multicolumn{1}{c}{\STAB{\rotatebox[origin=c]{90}{SVHN}}}
&\multicolumn{1}{c|}{\STAB{\rotatebox[origin=c]{90}{Sun397}}}
&\multicolumn{1}{c}{\STAB{\rotatebox[origin=c]{90}{Camelyon}}}
&\multicolumn{1}{c}{\STAB{\rotatebox[origin=c]{90}{EuroSAT}}}
&\multicolumn{1}{c}{\STAB{\rotatebox[origin=c]{90}{Resisc45}}}
&\multicolumn{1}{c|}{\STAB{\rotatebox[origin=c]{90}{Retinopathy}}}
&\multicolumn{1}{c}{\STAB{\rotatebox[origin=c]{90}{Clevr-Count}}}
&\multicolumn{1}{c}{\STAB{\rotatebox[origin=c]{90}{Clevr-Dist}}}
&\multicolumn{1}{c}{\STAB{\rotatebox[origin=c]{90}{DMLab}}}
&\multicolumn{1}{c}{\STAB{\rotatebox[origin=c]{90}{KITTI-Dist}}}
&\multicolumn{1}{c}{\STAB{\rotatebox[origin=c]{90}{dSpr-Loc}}}
&\multicolumn{1}{c}{\STAB{\rotatebox[origin=c]{90}{dSpr-Ori}}}
&\multicolumn{1}{c}{\STAB{\rotatebox[origin=c]{90}{sNORB-Azim}}}
&\multicolumn{1}{c|}{\STAB{\rotatebox[origin=c]{90}{sNORB-Ele}}}
&\multicolumn{1}{c}{\STAB{\rotatebox[origin=c]{90}{Average}}}\\
\specialrule{0em}{1pt}{1pt}
\hline
\specialrule{0em}{1pt}{1pt}
\multicolumn{22}{l}{\emph{Traditional Finetuning}}\\
\hline
\specialrule{0em}{1pt}{1pt}
Full&85.8&68.9&87.7&64.3&97.2&86.9&87.4&38.8&79.7&95.7&84.2&73.9&56.3&58.6&41.7&65.5&57.5&46.7&25.7&29.1&68.9 \\
Linear&0&64.4&85.0&63.2&97.0&86.3&36.6&51.0&78.5&87.5&68.5&74.0&34.3&30.6&33.2&55.4&12.5&20.0&9.6&19.2&57.6\\
\hline
\specialrule{0em}{1pt}{1pt}
\multicolumn{22}{l}{\emph{PETL methods}}\\
\hline
\specialrule{0em}{1pt}{1pt}
VPT&0.53&\bf78.8&90.8&65.8&98.0&88.3&78.1&49.6&81.8&\bf96.1&83.4&68.4&68.5&60.0&46.5&72.8&73.6&47.9&32.9&37.8&72.0 \\
Adapter&0.16&69.2&90.1&68.0&98.8&89.9&82.8&54.3&84.0&94.9&81.9&75.5&80.9&65.3&48.6&78.3&74.8&48.5&29.9&41.6&73.9 \\
AdaptFormer&0.16&70.8&91.2&70.5&99.1&90.9&86.6&54.8&83.0&95.8&84.4&76.3&81.9&64.3&49.3&80.3&76.3&45.7&31.7&41.1&74.7 \\
LoRA&0.29&67.1&91.4&69.4&98.8&90.4&85.3&54.0&\bf84.9&95.3&84.4&73.6&\bf82.9&\bf69.2&49.8&78.5&75.7&47.1&31.0&44.0&74.5
\\
NOAH&0.36&69.6&\bf92.7&70.2&99.1&90.4&86.1&53.7&84.4&95.4&83.9&\bf75.8&82.8&68.9&49.9&\bf81.7&81.8&48.3&32.8&44.2&75.5\\
\rowcolor{lightgray}Convpass$_\textit{attn}$&0.16&71.8&90.7&72.0&99.1&\bf91.0&89.9&54.2&85.2&95.6&83.4&74.8&79.9&67.0&50.3&79.9&84.3&\bf53.2&34.8&43.0&75.8
\\
\rowcolor{lightgray}Convpass&0.33&72.3&91.2&\bf72.2&\bf99.2&90.9&\bf91.3&\bf54.9&84.2&\bf96.1&\bf85.3&75.6&82.3&67.9&\bf51.3&80.0&\bf85.9&53.1&\bf36.4&\bf44.4&\bf76.6\\
\bottomrule[1.5pt]
\end{tabular}
}

\caption{\textbf{Full results on the VTAB-1K benchmark}. ``Average'' denotes the average results over three group-wise averages in Figure~\ref{fig:vtab}. ``\# params'' denotes the average number of trainable parameters \textbf{in backbones}. Convpass(-attn) achieves 12 SOTA results out of the 19 tasks among PETL methods.}
\label{tab:vtab}
\end{table*}
\begin{figure*}[t]
     \centering
         \includegraphics[width=0.92\textwidth]{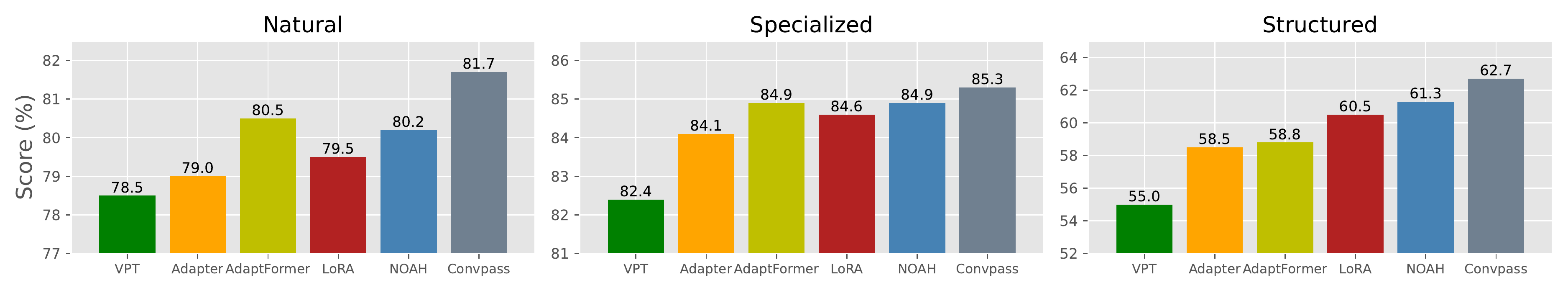}
         \caption{\textbf{Group-wise average results on VTAB-1K.} Convpass outperforms other baselines in all of the three groups.}
         \label{fig:vtab}
\end{figure*}

\subsection{Adapting ViT via Convolutional Bypasses}

Recent studies on modifying the architecture of ViT have verified that introducing convolution into ViT will improve the performance when training data is not adequate~\cite{vit,cvt}. Since the data of downstream tasks is usually limited even few-shot, we can also introduce convolution into the adaptation modules for PETL.

As illustrated in Figure~\ref{fig:cob}, a Convpass module consists of three convolutional layers: an $1\times1$ convolution reducing the channel, a $3\times3$ convolution with the same input and output channel, and an $1\times1$ convolution expanding the channel. Since ViT flattens the image into an 1D token sequence, we restore the 2D structure before convolution. The \texttt{[cls]} token serves as an individual image. The Convpass modules are placed parallel to the MHSA/MLP blocks, which can be formulated as
$$\boldsymbol{X}'\leftarrow\boldsymbol{X}+\textit{MHSA/MLP}(\textit{LN}(\boldsymbol{X}))+s\cdot\textit{Convpass}(\textit{LN}(\boldsymbol{X}))$$
where $s$ is a hyperparameter and $\textit{LN}$ is Layer Normalization~\cite{ln}. Note that the Convpass modules are similar to the \emph{residual bottleneck blocks} of ResNet~\cite{resnet}. If we ignore the MHSA/MLP blocks, the ViT will turn into a ResNet-like CNN.

From the unraveled view, we can find that in each transformer layer, besides the frozen paths, there are also trainable paths that only contain Convpass or contain both Convpass and MHSA which act as token-mixers. Therefore, the original transformer layers are converted to an ensemble of transformers, ResNet-like CNNs, and hybrid models. This design can help transfer learning from several perspective. First, since all the trainable paths contain Convpass modules, the finetuning process can benefit from the inherent visual inductive bias of CNN. Second, the 2D neighborhood structure of the $3\times3$ convolution focuses on local information, complementary to the MHSA that has global receptive field.

Convpass is storage-efficient. If the bottleneck channel size (i.e., the input \& output channel of the $3\times3$ convolution) is denoted as $h$, and the amount of ViT layers is $L$, the number of trainable parameters is $2L\times\left((2h+1)d+9h^2+2h\right)$. In view of $h<<d$ (e.g., $d=768, h=8$ in our experiments), this amount is $\mathcal{O}(Ld)$, which is negligible compared to ViT's $\mathcal{O}(Ld^2)$ parameters.

\section{Experiments}

\subsection{Transfer Learning on  VTAB-1K Benchmark}
First of all, our method is evaluated on the basic transfer learning scenario -- finetuning the pretrained models on various dowmstream tasks.

\subsubsection{Datasets}
To evaluate the performance on transfer learning of our methods, we use  VTAB-1K~\cite{vtab} as a benchmark.  VTAB-1K benchmark contains 19 image classification tasks from different fields, which can be roughly categorized into three groups: Natural, Specialized, and Structured. Each classification task only has 1,000 training samples, which are split into a training set (800) and a validation set (200) during hyperparameter search. The reported results are produced by evaluating the model trained on all the 1,000 training samples on test set.
\begin{figure*}[t]
     \centering
         \includegraphics[width=0.98\textwidth]{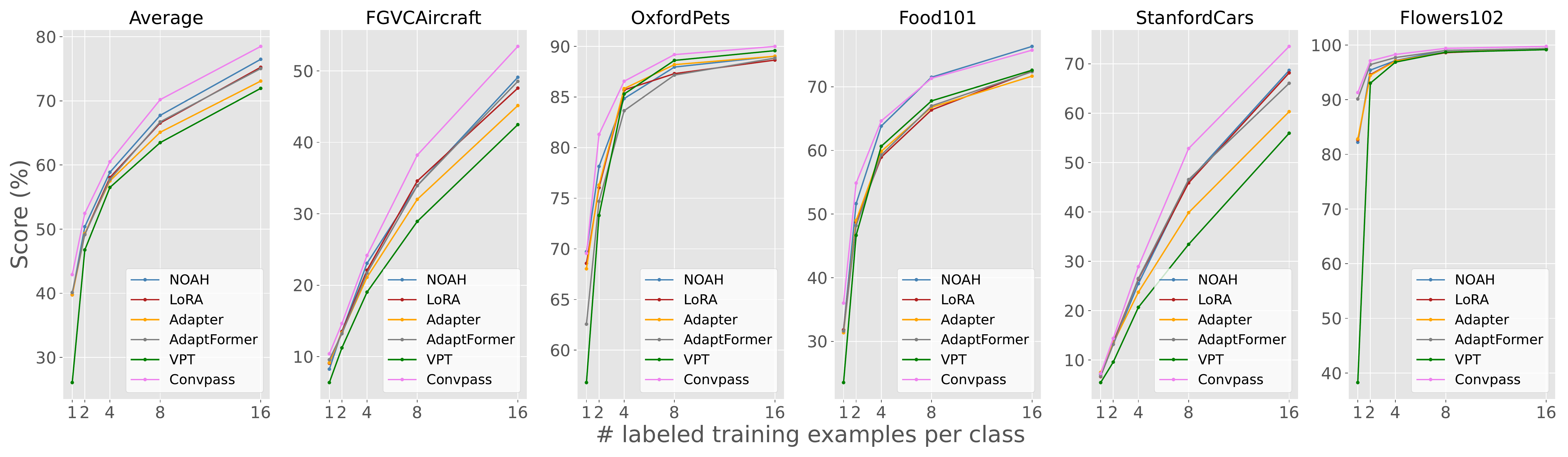}
         \caption{\textbf{Results of few-shot learning on five fine-grained visual recognition datasets. } Convpass outperforms other baselines on average results.}
         \label{fig:fs}
\end{figure*}

\subsubsection{Baselines}
We compare our method with two traditional finetuning methods: \textbf{Full} finetuning, which optimizes all parameters end-to-end; \textbf{Linear} evaluation, which freezes the pretrained backbone and only learns a classification head; as well as four PETL methods: \textbf{VPT}, \textbf{Adapter}, \textbf{AdaptFormer}, \textbf{LoRA}, and \textbf{NOAH}. For our method \textbf{Convpass}, we also report a simplified variant: \textbf{Convpass$_\textit{attn}$}, which only inserts the Convpass modules alongside the MHSA blocks. Note that VPT, Adapter, LoRA, and Convpass only contain one type of PETL module, and the network architecture is the same for all tasks; while NOAH focuses on architecture search to combine other existing PETL modules, resulting in a dynamic network architecture. 


\subsubsection{Setup}We use a ViT-B/16~\cite{vit} supervisedly pretrained on ImageNet-21K~\cite{imagenet} for al methods. The networks are finetuned for 100 epochs except for NOAH, which also trains a supernet for another 500 epochs. The hidden dimension $h$ of Adapter, AdaptFormer and Convpass, as well as the rank $r$ of LoRA are all set to 8. The prompt length $l$ of VPT follows the best recipe in its original paper. The hyperparameter $s$ of Convpass and AdaptFormer is roughly searched from \{0.01, 0.1, 1, 10, 100\}. In this setting, Adapter, AdaptFormer and Convpass$_\textit{attn}$ have similar numbers of trainable parameters, while the Convpass's trainable parameters are slightly more than LoRA's but fewer than VPT's and NOAH's. Other hyperparameters are listed in the Appendix.

\subsubsection{Results}As shown in Table~\ref{tab:vtab}, Convpass$_\textit{attn}$ outperforms its counterpart Adapter and AdaptFormer on 16 and 10 out of the 19 tasks, while Convpass outperforms its counterparts LoRA and NOAH on 15 and 13 tasks, respectively. Although using fewer parameters, Convpass still performs better than VPT on 17 tasks. All the PETL methods are better than full finetuning overall. Because of the variety of tasks, no one method achieves SOTA on all tasks at once, but Convpass achieves the best average performance, $1.1\%$ higher than the previous SOTA PETL methods, NOAH. It is worth noting that Convpass$_\textit{attn}$ also has better average results than NOAH with only half as many parameters as NOAH. Moreover, since NOAH need to train an additional large supernet for architecture search, Convpass is also superior to NOAH in terms of training efficiency.

Figure~\ref{fig:vtab} shows that Convpass has the best performance in all the three groups of VTAB, indicating that Convpass specializes in visual tasks from various domains. The superiority of Convpass is significant in the Natural and Structured groups. But in the Specialized group, Convpass does not remarkably outperform NOAH and AdaptFormer.

\subsection{Few-Shot Learning}

Few-shot learning is a common scenario when the data of downstream tasks is hard to obtain, and there are only a few training samples for each task that can be utilized.

\subsubsection{Datasets}We use five fine-gained datasets to evaluate the performance of our methods on few-shot learning: \textbf{FGVC-Aircraft}~\cite{aircraft}, \textbf{Oxford-Pets}~\cite{pets}, \textbf{Food-101}~\cite{food}, \textbf{Stanford Cars}~\cite{car}, and \textbf{Oxford-Flowers102}~\cite{flower}. We conduct experiments on 1, 2, 4, 8, and 16 shot settings. The results are averaged over three runs with different seeds. The experimental setup and baselines are the same as for VTAB-1K.

\subsubsection{Results}As shown in Figure~\ref{fig:fs}, the average results of Convpass are all higher than the other baselines across the five settings. On FGVC-Aircraft and Stanford Cars, the advantages of Convpass are highlighted. On simpler Oxford-Pets and  Oxford-Flowers102, all the methods have similar performance, while Convpass is still in the lead. On Food-101, Convpass slightly  underperforms NOAH in the 16-shot case, but the trend is reversed when the number of training data gets smaller. These results demonstrate that the introduced inductive bias of Convpass enhances ViT's capability to learn in the low-data regime.

\subsection{Domain Generalization}

Besides vision models, PETL has been studied in the field of vision-language models as well. Considering the outstanding domain generalization property of vision-language models, we also evaluate the performance of our method  under domain shift when applied to vision-language models.

\subsubsection{Datasets}In domain generalization experiments, the models are trained on the source domain, and tested on both the source and target domain. We use \textbf{ImageNet-1K}~\cite{imagenet} as the source domain, where each class contains 16 training samples. The target domains include: \textbf{ImageNet-V2}~\cite{inv2}, which is a new ImageNet test set collected with the original labelling protocol; \textbf{ImageNet-Sketch}~\cite{insk}, which consists of sketch images of the 1,000 ImageNet classes; \textbf{ImageNet-A}~\cite{ina}, which contains real-world adversarial samples of 200 of the ImageNet classes; \textbf{ImageNet-R}~\cite{inr}, which is composed of renditions of 200 ImageNet classes.

\begin{table}[t]
\centering
\setlength{\tabcolsep}{4.5pt}

\scalebox{0.9}{
\begin{tabular}{p{2cm}<{\centering}p{1.3cm}<{\centering}p{0.0cm}<{\centering}p{0.8cm}<{\centering}p{0.8cm}<{\centering}p{0.8cm}<{\centering}p{0.8cm}<{\centering}}
\toprule[1.5pt]
\multirow{2}{*}{Method} & Source && \multicolumn{4}{c}{Target}\\
\cline{2-2}\cline{4-7}
\specialrule{0em}{1pt}{1pt}
&ImageNet&&-V2&-S\tiny ketch&-A&-R\\\hline
\specialrule{0em}{1pt}{1pt}
ZS CLIP&66.73&&60.83&46.15&47.77&73.96\\\specialrule{0em}{1pt}{1pt}
LP CLIP&65.85&&56.26&34.77&35.68&58.43\\\specialrule{0em}{1pt}{1pt}
CoOp&71.51&&64.20&47.99&49.71&75.21\\\specialrule{0em}{1pt}{1pt}
CoCoOp&71.02&&64.07&48.75&\bf50.63&76.18\\\specialrule{0em}{1pt}{1pt}
\small Tip-Adapter-F&73.41&&65.39&48.58&49.23&77.54\\\specialrule{0em}{1pt}{1pt}

\rowcolor{lightgray}\small Convpass$_\textit{CLIP}$&\bf74.23&&\bf66.61&\bf49.10&49.27&\bf78.17\\
\bottomrule[1.5pt]
\end{tabular}
}
\caption{\textbf{Results of 16-shot ImageNet classification and domain generalization on CLIP. }We report top-1 accuracy. Convpass$_\textit{CLIP}$ outperforms the baselines on source domain and three of the four target domains.}
\label{tab:dg}
\end{table}

\subsubsection{Baselines} The CLIP~\cite{clip} model consists of an image encoder and a text encoder, which are pretrained via contrastive learning on image-text pairs. Our method is compared with the following baselines: \textbf{Zero-Shot (ZS) CLIP} uses prompted label texts (e.g., ``A photo of \texttt{<class name>}.'') as the text encoder inputs, and classifies the images based on cosine similarity between image and text features ; \textbf{Linear Probe (LP) CLIP} discards the text encoder and learns a linear classification head for image encoder; \textbf{CoOp}~\cite{coop} makes use of trainable vectors as prompts of labels; \textbf{CoCoOp}~\cite{cocoop} learns a meta-net to generate prompts of labels from images; \textbf{Tip-Adapter-F}~\cite{tip} caches features of training data to initialize an adapter after the image encoder. Note that CoOp, CoCoOp, and Tip-Adapter-F are PETL methods designed for CLIP specifically.

To apply our methods to CLIP, we make the following modifications. \textbf{First}, we insert Convpass modules into the image encoder only, while the text encoder stays unchanged. \textbf{Second}, we add a FC layer as classification head of the image encoder, whose bias is zero-initialized and whose weight is initialized with encoded prompted label texts of all classes (just as in ZS CLIP). Then, the text encoder is discarded, and only the Convpass modules and head are finetuned. We call this ported PETL method \textbf{Convpass$_\textit{CLIP}$}.

\subsubsection{Setup}In our experiments, all methods use a ViT-B/16 as the image encoder, and a BERT-like~\cite{bert} model as the text encoder. For our methods, we train the Convpass modules and classification heads for 50 epochs. Other hyperparameters are listed in the Appendix.

\subsubsection{Results}The results are shown in Table~\ref{tab:dg}. Our method, though not designed for CLIP, still outperforms the baselines tailored for CLIP on the source domain. On three out of the four target domains, Convpass$_\textit{CLIP}$ also achieves SOTA performance. On ImageNet-A, Convpass$_\textit{CLIP}$ performs a bit poorly, which is probably because the ImageNet-A dataset is collected by selecting samples misclassified by ResNet. Since Convpass modules are ResNet-style blocks, they may be more easily misled by these samples as well. Overall, the results prove that Convpass$_\textit{CLIP}$ is robust under domain shift. 

\begin{table}[t]
\centering
\scalebox{0.9}{
\begin{tabular}{p{1.9cm}<{\centering}p{1.3cm}<{\centering}p{0.8cm}<{\centering}p{0.8cm}<{\centering}p{0.8cm}<{\centering}p{0.8cm}<{\centering}}
\toprule[1.5pt]
\specialrule{0em}{1pt}{1pt}
Model&Method&Avg.&Nat.&Spe.&Str.\\\hline
\specialrule{0em}{1pt}{1pt}
ConvNeXt-B&Full&74.0&78.0&83.7&60.4\\\specialrule{0em}{1pt}{1pt}
ConvNeXt-B&Linear&63.6&74.5&81.5&34.8\\\specialrule{0em}{1pt}{1pt}
\hline\specialrule{0em}{1pt}{1pt}

Swin-B&Full&75.0&79.2&86.2&59.7\\\specialrule{0em}{1pt}{1pt}
Swin-B&Linear&62.6&73.5&80.8&33.5\\\specialrule{0em}{1pt}{1pt}
Swin-B&VPT&71.6&76.8&84.5&53.4\\\specialrule{0em}{1pt}{1pt}
\rowcolor{lightgray}Swin-B&Convpass&\bf78.1&\bf83.1&\bf87.2&\bf64.1\\\specialrule{0em}{1pt}{1pt}
\hline\specialrule{0em}{1pt}{1pt}

ViT-B/16&Full&68.9&75.9&83.4&47.6\\\specialrule{0em}{1pt}{1pt}
ViT-B/16&Linear&57.6&68.9&77.2&26.8\\\specialrule{0em}{1pt}{1pt}
ViT-B/16&VPT&72.0&78.5&82.4&55.0\\\specialrule{0em}{1pt}{1pt}
\rowcolor{lightgray}ViT-B/16&Convpass&\bf76.6&\bf81.7&\bf85.3&\bf62.7\\
\bottomrule[1.5pt]
\end{tabular}
}
\caption{\textbf{Results on  VTAB-1K. }ConvNeXt-B and Swin-B have inherent inductive bias for vision, while ViT introduces such inductive bias via Convpass during finetuning. Avg.: Average, Nat.: Natural, Spe.: Specialized, Str.: Structured.}
\label{tab:fa3}
\end{table}

\subsection{Further Analyses}

\subsubsection{Comparison with Other Backbones}
One of the motivations for designing Convpass is to introduce visual inductive bias to ViT during finetuning. However, since there are also ViT variants (e.g., Swin Transformer~\cite{swin}) which have already incorporated visual inductive bias into their model designs, finetuning on these models can naturally benefit from such prior knowledge. Then a question arises: \emph{Can the models that acquire inductive bias during finetuning outperform these models that have innately hard-coded inductive bias?}

We conduct comparisons among the three backbone models: \textbf{ViT-B/16}, \textbf{Swin-B}~\cite{swin}, and a SOTA CNN \textbf{ConvNeXt-B}~\cite{convnext}. All of them are pretrained on ImageNet-21K and have a similar size. As the results shown in Table~\ref{tab:fa3}, when using traditional transfer learning methods (Full and Linear), Swin-B and ConvNeXt-B perform significantly better than ViT-B/16 as expected, which indicates the pivotal role of visual inductive bias during finetuning. However, when equipped with Convpass, the average performance of ViT-B/16 overtakes fully finetuned Swin-B and ConvNeXt-B. These observations suggest that Convpass has the powerful capability  to complement the missing inductive bias for downstream transfer tasks.

\begin{figure*}[t]
     \centering
     \begin{minipage}[t]{0.82\textwidth}
     \centering
     \begin{subfigure}[b]{0.20\textwidth}
         \centering
         \includegraphics[width=\textwidth]{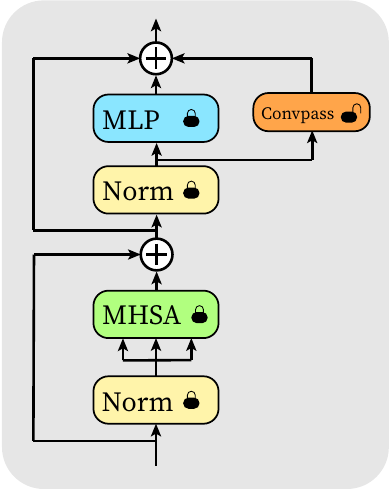}
         \caption{Convpass$_\textit{mlp}$}
     \end{subfigure}
     \hfill
     \begin{subfigure}[b]{0.20\textwidth}
         \centering
         \includegraphics[width=\textwidth]{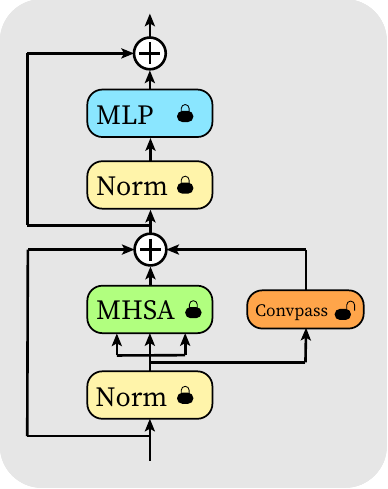}
         \caption{Convpass$_\textit{attn}$}
     \end{subfigure}
     \hfill
     \begin{subfigure}[b]{0.20\textwidth}
         \centering
         \includegraphics[width=\textwidth]{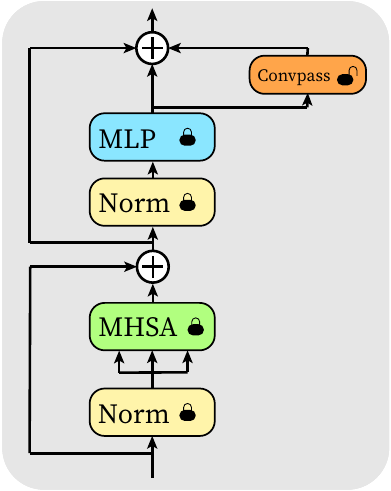}
         \caption{Seq-Convpass$_\textit{mlp}$}
     \end{subfigure}
          \hfill
     \begin{subfigure}[b]{0.20\textwidth}
         \centering
         \includegraphics[width=\textwidth]{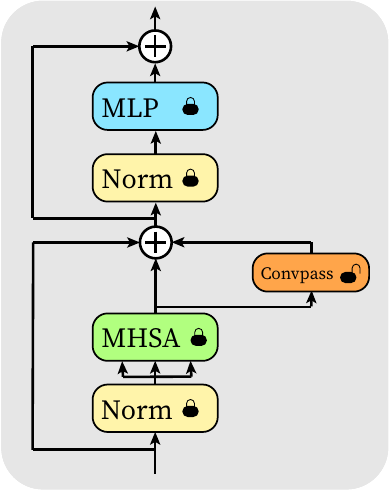}
         \caption{Seq-Convpass$_\textit{attn}$}
     \end{subfigure}
        \caption{\textbf{Four ways to insert a  Convpass module into ViT. }}
        \label{fig:fa4}
    \end{minipage}
    
\end{figure*}

Moreover, we also apply Convpass to Swin. Similarly, Convpass modules bypass the W-MHSA/SW-MHSA/MLP blocks of Swin. As shown in Table~\ref{tab:fa3}, the advantage of Convpass over full finetuning still holds on Swin, but VPT is no longer competitive. This observation demonstrates that Convpass is a reliable PETL method performing constantly well on various backbone networks. From the comparison between Swin and ViT we also find that the improvement made by Convpass diminishes on Swin. This is also expected because the demand for supplementing visual inductive on Swin is not as pressing as on ViT.

\begin{table}[t]
\centering
\scalebox{1}{
\begin{tabular}{p{3cm}<{\centering}p{0.8cm}<{\centering}p{0.8cm}<{\centering}p{0.8cm}<{\centering}p{0.8cm}<{\centering}}
\toprule[1.5pt]
\specialrule{0em}{1pt}{1pt}
Method&Avg.&Nat.&Spe.&Str.\\
\hline
\specialrule{0em}{1pt}{1pt}

Seq-Convpass$_\textit{mlp}$&74.5&80.0&83.6&59.9\\\specialrule{0em}{1pt}{1pt}
Seq-Convpass$_\textit{attn}$&74.9&80.6&84.1&60.0\\\specialrule{0em}{1pt}{1pt}
Convpass$_\textit{mlp}$&75.4&80.3&84.5&61.2\\\specialrule{0em}{1pt}{1pt}
Convpass$_\textit{attn}$&\bf75.8&\bf81.2&\bf84.7&\bf61.5\\

\bottomrule[1.5pt]
\end{tabular}
}
\caption{\textbf{Results on  VTAB-1K. }We find that (i) parallel is superior to sequential, and (ii) alongside MHSA is superior to alongside MLP. }
\label{tab:fa4}
\end{table}

\begin{table}[t]
\centering
\scalebox{1}{
\begin{tabular}{p{1.2cm}<{\centering}p{1.2cm}<{\centering}p{0.8cm}<{\centering}p{0.8cm}<{\centering}p{0.8cm}<{\centering}p{0.8cm}<{\centering}}
\toprule[1.5pt]
\specialrule{0em}{1pt}{1pt}
@MLP&@MHSA&Avg.&Nat.&Spe.&Str.\\
\hline\specialrule{0em}{1pt}{1pt}
1$\times$1&1$\times$1&75.1&81.1&84.8&59.5\\\specialrule{0em}{1pt}{1pt}
3$\times$3&1$\times$1&75.8&81.2&84.4&61.7\\\specialrule{0em}{1pt}{1pt}
1$\times$1&3$\times$3&75.8&81.3&84.4&61.5\\\specialrule{0em}{1pt}{1pt}
\rowcolor{lightgray}3$\times$3&3$\times$3&\bf76.6&\bf81.7&\bf85.3&\bf62.7\\

\bottomrule[1.5pt]
\end{tabular}
}
\caption{\textbf{Results on  VTAB-1K. }``1$\times$1 @MLP'' means the 3$\times$3 convolutions in Convpass modules alongside the MLP blocks are replaced with 1$\times$1 convolutions. We find that vision-oriented is superior to language-oriented. }
\label{tab:fa5}
\end{table}

\subsubsection{Where to Place the Convpass Modules}
Our Convpass modules are parallel to the MHSA/MLP blocks, but there is another choice: insert the modules after the MHSA/MLP blocks in a sequential way like Adapter. To figure out what is the optimal way to place the Convpass modules, we consider four forms when only one Convpass module is inserted in each ViT layer , as illustrated in Figure~\ref{fig:fa4}. \textbf{Convpass$_\textit{mlp}$} and \textbf{Convpass$_\textit{attn}$} are parallel Convpass modules alongside the MLP and MHSA blocks, while \textbf{Seq-Convpass$_\textit{mlp}$} and \textbf{Seq-Convpass$_\textit{attn}$} follow the MLP and MHSA blocks, respectively. 

As shown in Table~\ref{tab:fa4}, we evaluate these designs on  VTAB-1K, and find the following. \textbf{First}, the parallel designs are better than their sequential counterparts. From Figure~\ref{fig:unravel}, we know that the sequential modules add longer paths to the model, which are relatively harder to optimize with a small amount of downstream data. On the contrary, the parallel Convpass serve as shortcuts for better gradient propagation, and 
introduce fully-convolutional ResNet-like paths that do not exist in sequential designs. \textbf{Second}, we also find that placing the Convpass modules beside/after MHSA blocks is better than beside/after MLP blocks.  Since Convpass$_\textit{attn}$ and Convpass$_\textit{mlp}$ are the best two designs, our Convpass is composed of them, i.e., placing Convpass modules alongside both MHSA and MLP blocks in parallel.

\subsubsection{Vision-Oriented vs. Language-Oriented}
Finally, we conduct an ablation study on the vision-oriented idea. As shown in Table~\ref{tab:fa5}, we replace the 3$\times$3 convolutions in Convpass modules alongside the MLP and/or MHSA with 1$\times$1 convolutions, yielding four different designs. The bottom row is exactly Compass. Replacing the 3$\times$3 convolution in a Convpass module means the module will lose its capacity as a token mixer, degrading into a language-oriented adaptation module similar to Adapter.

The results show that, whether we replace the 3$\times$3 convolutions of Convpass modules alongside MLP or alongside MHSA, the performance on Natural, Specialized, and Structured tasks will all degrade. If all 3$\times$3 convolutions are replaced, the model will perform rather poorly on Structured. Since the Structured tasks are about obtaining the structure of a scene (e.g., object counting or 3D depth prediction), they fairly differ from the pretraining tasks (i.e., ImageNet classification) and require more modifications to the pretrained token mixer. Therefore, the Structured tasks are more complicated and the superiority of vision-oriented modules is highlighted.
In summary, the language-oriented ablation models perform worse than the vision-oriented Convpass, supporting our standpoint.

\section{Conclusion}
In this paper, we point out that current PETL methods used in ViT lack inductive bias for visual tasks, which potentially degrades the performance on downstream finetuning. For this reason, we propose Convpass, a vision-oriented PETL method that employs trainable convolotional bypasses to adapt pretrained ViT to downstream tasks. Experimental results on  VTAB-1K benchmark and few-shot learning show that Convpass outperforms other PETL methods and owns remarkable domain generalization property. Our simple but effective method reveals the importance of considering the characteristics of visual tasks when designing ViT-based PETL methods, which lights a promising direction for future work.

\onecolumn
\section*{Appendix}
\setcounter{section}{0}

\renewcommand\thesection{\Alph{section}}
\section{Datasets}
See Table~\ref{tab:dataset}. Since the test set of ImageNet-1K has not been released, we report its validation results in our experiments.
\begin{table*}[h]

\centering
\scalebox{0.9}{
\begin{tabular}{clcccc}
\toprule[1.5pt]
                          & Dataset              & \# Classes & Train                    & Val   & Test            \\ \midrule
                          \multicolumn{6}{c}{VTAB-1K~\cite{vtab}}\\\midrule
\multirow{7}{*}{Natural} & CIFAR100~\cite{krizhevsky2009learning}             & 100       & \multirow{7}{*}{800/1,000}                 & \multirow{7}{*}{200}   & 10,000           \\
                          & Caltech101~\cite{fei2004learning}           & 102       &                 &    & 6,084            \\
                          & DTD~\cite{cimpoi14describing}                 & 47        &                 &    & 1,880            \\
                          & Oxford-Flowers102~\cite{flower}    & 102       &                 &    & 6,149            \\
                          & Oxford-Pets~\cite{pets}          & 37        &                 &    & 3,669     \\
                          & SVHN~\cite{netzer2011reading}                 & 10        &                 &    & 26,032            \\
                          & Sun397~\cite{xiao2010sun}               & 397       &                 &    & 21,750           \\
\midrule\multirow{4}{*}{Specialized}                          & Patch Camelyon~\cite{Veeling2018qh}       & 2         &      \multirow{4}{*}{800/1,000}                 & \multirow{4}{*}{200}                & 32,768           \\
                          & EuroSAT~\cite{helber2019eurosat}              & 10        &                 &    & 5,400            \\
                          & Resisc45~\cite{cheng2017remote}             & 45        &                 &    & 6,300            \\
                          & Retinopathy~\cite{kaggle2015retinopathy}         & 5         &                 &    & 42,670           \\
 \midrule\multirow{8}{*}{Structured}                         & Clevr/count~\cite{johnson2017clevr}          & 8         &      \multirow{8}{*}{800/1,000}                 & \multirow{8}{*}{200}                  & 15,000       \\
                          & Clevr/distance~\cite{johnson2017clevr}       & 6         &                 &    & 15,000     \\
                          & DMLab~\cite{beattie2016deepmind}                & 6         &                 &    & 22,735           \\
                          & KITTI-Dist~\cite{geiger2013vision}           & 4         &                 &    & 711   \\
                          & dSprites/location~\cite{matthey2017dsprites}    & 16        &                 &    & 73,728           \\
                          & dSprites/orientation~\cite{matthey2017dsprites} & 16        &                 &    & 73,728           \\
                          & SmallNORB/azimuth~\cite{lecun2004learning}    & 18        &                 &    & 12,150     \\
                          & SmallNORB/elevation~\cite{lecun2004learning}  & 18         &                 &    & 12,150     \\ \midrule
                           \multicolumn{6}{c}{Few-shot learning}\\\midrule
 & Food-101~\cite{food}             & 101       & \multirow{5}{*}{1/2/4/8/16 per class} & 20,200 & 30,300            \\
                          & Stanford Cars~\cite{car}        & 196       &  & 1,635  & 8,041       \\
                          & Oxford-Flowers102~\cite{flower}    & 102       &  & 1,633  & 2,463             \\
                          & FGVC-Aircraft~\cite{fgvc}        & 100       &  & 3,333  & 3,333       \\
                          & Oxford-Pets~\cite{pets}          & 37        &  & 736   & 3,669    \\\midrule
                           \multicolumn{6}{c}{Domain generalization}\\\midrule
 & ImageNet-1K~\cite{imagenet}             & 1,000       & 16 per class &  50,000 & N/A             \\
                          & ImageNet-V2~\cite{inv2}        & 1,000       & N/A & N/A   & 10,000       \\
                          & ImageNet-Sketch~\cite{insk}    & 1,000       &N/A   & N/A   & 50,889             \\
                          & ImageNet-A~\cite{ina}        & 200       & N/A  & N/A   & 7,500       \\
                          & ImageNet-R~\cite{inr}          & 200        & N/A  & N/A    & 30,000    \\\bottomrule[1.5pt]
\end{tabular}}
\caption{\textbf{Statistics of used datasets.}
}
\label{tab:dataset}
\end{table*}

\section{Experimental Details}
\subsection{Pretrained Backbones}
See Table~\ref{tab:pt}.

\begin{table*}[h]

\centering
\begin{minipage}{\linewidth}
\centering

\scalebox{1}{

\begin{tabular}{lccc}
\toprule[1.5pt]
Model&Pretraining Dataset&Size (M)&Pretrained Weights
\\\midrule
ViT-B/16~\cite{vit}&ImageNet-21K&85.8&\footnote{https://storage.googleapis.com/vit\_models/imagenet21K/ViT-B\_16.npz}{checkpoint}
\\
Swin-B~\cite{swin}&ImageNet-21K&86.7&\footnote{https://github.com/SwinTransformer/storage/releases/download/v1.0.0/swin\_base\_patch4\_window7\_224\_22k.pth}{checkpoint}
\\
ConvNeXt-B~\cite{convnext}&ImageNet-21K&87.6&\footnote{https://dl.fbaipublicfiles.com/convnext/convnext\_base\_22k\_224.pth}{checkpoint}
\\
CLIP ViT-B/16~\cite{clip}&WebImageText&85.8 (image encoder)&\footnote{https://openaipublic.azureedge.net/clip/models/5806e77cd80f8b59890b7e101eabd078d9fb84e6937f9e85e4ecb61988df416f/ViT-B-16.pt}{checkpoint}
\\
 \bottomrule[1.5pt]
\end{tabular}
}
\end{minipage}

\caption{\textbf{Pretrained backbones.}
}
\label{tab:pt}
\end{table*}

\subsection{Code Implementation}
We use \footnote{https://pytorch.org/}{\emph{PyTorch}} to implement all experiments on NVIDIA RTX3090 GPUs. The models are implemented based on \footnote{https://rwightman.github.io/pytorch-image-models/}{\emph{timm}}.

\subsection{Data Augmentation}
\subsubsection{VTAB-1K} Following \cite{vpt,noah}, we resize the images to $224\times224$, and then normalize them with ImageNet's mean and standard deviation.
\subsubsection{Few-shot learning} Following \cite{noah}, for training samples, we use color-jitter and RandAugmentation; for validation/test samples, we resize them to $256\times256$, crop them to $224\times224$  at the center, and then normalize them with ImageNet's mean and standard deviation.
\subsubsection{Domain generalization} Following \cite{cocoop}, for training samples, we randomly resize and crop them to $224\times224$, and then implement random horizontal flip; for validation/test samples, we resize them to  $224\times224$. All samples are finally normalized with ImageNet's mean and standard deviation.

\subsection{Hyperparameters}
$s$ is searched from \{0.01, 0.1, 1, 10, 100\}. See Table~\ref{tab:hyper} for other hyperparameters.
\begin{table*}[h]

\centering
\scalebox{0.9}{
\begin{tabular}{cccccccc}
\toprule[1.5pt]
 &optimizer&batch size&learning rate&weight decay&\# epochs&lr decay&\# warm-up epochs\\ \midrule
 VTAB-1K&AdamW&64&1e-3&1e-4&100&cosine&10\\
 Few-shot learning&AdamW&64&5e-3&1e-4&100&cosine&10\\
 Domain generalization&AdamW&64&1e-5&0&50&cosine&0\\
 \bottomrule[1.5pt]
\end{tabular}}
\caption{\textbf{Hyperparameters.}
}
\label{tab:hyper}
\end{table*}

\subsection{Results of Adapter}
We notice that it is reported by \citet{vpt} that Adapter significantly underperforms VPT. This is because their implementation uses zero-initialization for weights of both the two FC layers in Adapter, which blocks the backpropagation of gradient. Therefore, we follow \citet{noah} using Xavier-initialization for weights and zero-initialization for biases, and report results that Adapter outperforms VPT.

\twocolumn
\bibliography{aaai23.bib}

\end{document}